\pdfoutput=1
\documentclass{article}

    \PassOptionsToPackage{numbers, compress}{natbib}
\usepackage{amsmath}
\usepackage[final]{neurips_2022}

\usepackage[utf8]{inputenc} % allow utf-8 input
\usepackage[T1]{fontenc}    % use 8-bit T1 fonts

\usepackage{url}            % simple URL typesetting
\usepackage{booktabs}       % professional-quality tables
\usepackage{amsfonts}       % blackboard math symbols
\usepackage{nicefrac}       % compact symbols for 1/2, etc.
\usepackage{microtype}      % microtypography
\usepackage{xcolor}         % colors
\usepackage{graphicx}
\usepackage{hyperref}       % hyperlinks

\title{Self-recovery of memory via generative replay}

\author{%
  Zhenglong Zhou, Geshi Yeung, \& Anna C. Schapiro \\
  University of Pennsylvania, Department of Psychology\\
  \texttt{\{zzhou34, geshi, aschapir\}@sas.upenn.edu} \\
}

\begin{document}

\maketitle

\begin{abstract}
  A remarkable capacity of the brain is its ability to autonomously reorganize memories during offline periods. Memory replay, a mechanism hypothesized to underlie biological offline learning, has inspired offline methods for reducing forgetting in artificial neural networks in continual learning settings. A memory-efficient and neurally-plausible method is generative replay, which achieves state of the art performance on continual learning benchmarks. However, unlike the brain, standard generative replay does not self-reorganize memories when trained offline on its own replay samples. We propose a novel architecture that augments generative replay with an adaptive, brain-like capacity to autonomously recover memories. We demonstrate this capacity of the architecture across several continual learning tasks and environments. 
\end{abstract}

\section{Introduction}

The brain strengthens and reorganizes its own memories during offline periods, especially sleep \cite{landmann_reorganisation_2014, rasch_about_2013, stickgold_sleep-dependent_2005}. A mechanism hypothesized to underlie this capacity is the replay of neural activity associated with awake experience \cite{foster_replay_2017}, which has been identified across species and shown to facilitate memory \cite{foster_replay_2017, liu_decoding_2022}. Theories posit that re-organizing memories offline is the brain's key to continually learning new tasks without disrupting existing knowledge across a lifetime \cite{mcclelland_why_1995}. 

\begin{figure}[!ht]
\centering
\includegraphics[scale=0.40]{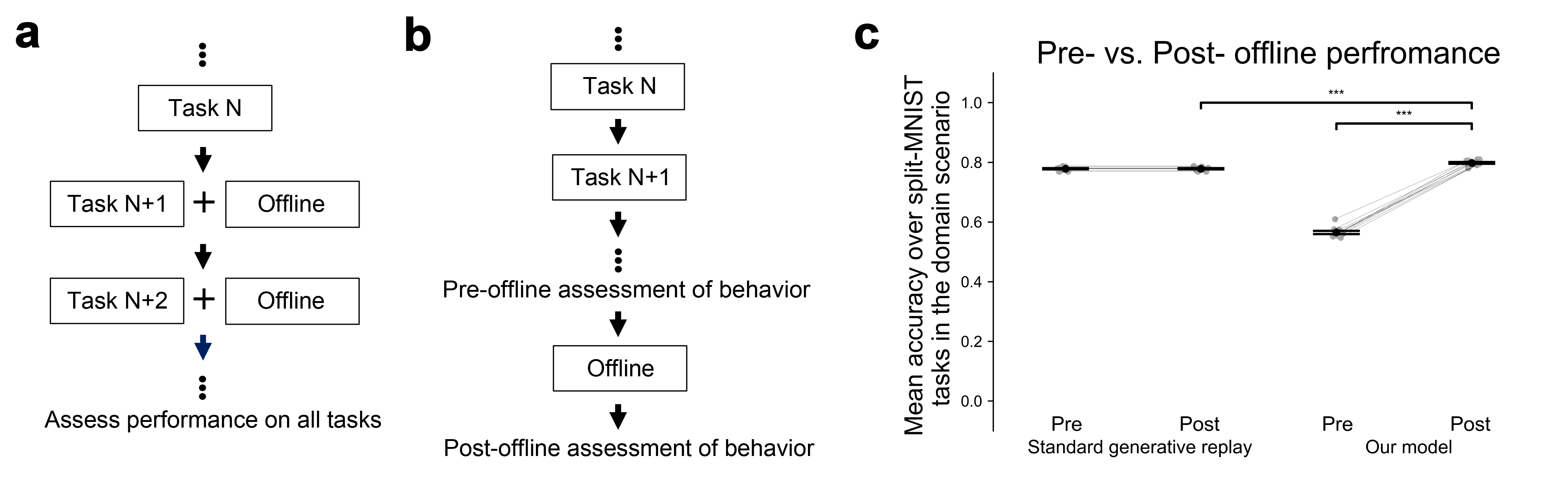}
% \captionsetup{width=0.85\linewidth,font=footnotesize}
\caption{Differences in the assessment of artificial and biological offline learning. \textbf{a}. In standard continual learning settings, offline methods are assessed on their ability to help preserve memories as ANNs sequentially learn a set of tasks. \textbf{b}. In neuroscience and psychology, studies of offline learning examine how behavior changes across an offline period in the absence of external inputs. \textbf{c}. When assessed as in empirical studies, standard generative replay does not benefit behavior, whereas our model shows an improvement in performance through offline learning. Each dot shows the performance of a model initialized with a unique random seed. The pre- and post- offline accuracy of each model instantiation is connected by a grey line. Error bars represent $\pm$ 1 SEM. *** indicates p<0.001.
}
\label{fig:examples}
\end{figure}

In artificial neural networks (ANNs), learning new tasks without forgetting previous tasks (i.e., continual learning) remains a core challenge: In non-stationary settings where the data from previous tasks are no longer accessible when learning a new task, incorporating new information can catastrophically damage existing knowledge \cite{mccloskey_catastrophic_1989, french_catastrophic_1999}. Inspired by neuroscience, an approach to continual learning is to replay data that reflect prior tasks when learning a new task \cite{mnih_human-level_2015, shin_continual_2017, van_de_ven_brain-inspired_2020, chaudhry_tiny_2019, hayes_replay_2021, kemker_fearnet_2017, li_learning_2017, rebuffi_icarl_2017, robins_catastrophic_1995, rolnick_experience_2019, schwarz_progress_2018,ratcliff_connectionist_1990} Replaying memories in a way that comprehensively reflects prior experience allows information across temporally-separated experiences to be integrated gracefully. The most straightforward implementation of this approach stores veridical inputs of past tasks in a memory buffer, from which data can be drawn for subsequent replay \cite{rebuffi_icarl_2017, rolnick_experience_2019, chaudhry_tiny_2019}. However, replaying veridical inputs is unlikely to scale in real-world scenarios (e.g., when a general-purpose robot has to learn a large number of tasks across a lifetime) and is biologically implausible \cite{krause_large_2022, stella_hippocampal_2019, carr_hippocampal_2011}. A promising extension of the approach is generative replay \cite{shin_continual_2017, van_de_ven_brain-inspired_2020}, in which a generative model learns to reconstruct previous inputs, avoiding storage of exact inputs. This form of replay is considered more neurally-plausible \cite{van_de_ven_brain-inspired_2020}; the brain does not store exact copies of experience, and biological replay resembles noisy samples from a generative model of the environment \cite{krause_large_2022, stella_hippocampal_2019}. On continual learning benchmarks in which a set of tasks have to be learned sequentially, generative replay shows state of the art performance \cite{van_de_ven_three_2019, van_de_ven_brain-inspired_2020}.

Thus, generative replay appears to be a useful mechanism for both biological and artificial learning. However, despite its similarity to neural replay, generative replay lacks a brain-like capacity to adaptively self-reorganize memories offline. In this work, we first highlight a crucial difference between how biological and artificial replay are typically assessed. We point out how, when assessed in a way similar to empirical studies of offline processing, standard generative replay does not benefit behavior. We propose a novel architecture that endows generative replay with a brain-like capacity of self-reorganization: The ability to self-repair damaged memories through offline processing. We demonstrate this capacity of the model on MNIST \cite{lecun_gradient-based_1998} and CIFAR-100 \cite{krizhevsky_learning_2009}.

\section{Standard generative replay does not self-reorganize offline}

Psychology and neuroscience research has investigated the ways in which the brain autonomously reorganizes memories offline to enhance behavior. This work suggest that the offline brain can go beyond what's learned during wakefulness, giving rise to useful changes in behavior through spontaneous processes \cite{landmann_reorganisation_2014, rasch_about_2013, stickgold_sleep-dependent_2005}. These changes are indexed by differences in behavior before versus after an offline period (Fig. 1b). For example, after learning a sequence of tasks, a period of sleep can repair memories that have been damaged due to interference in a preceding wakeful period \cite{mcdevitt_rem_2015, mednick_restorative_2002, baran_rem-dependent_2010, drosopoulos_sleeps_2007} — an adaptive capacity that biologically detailed models of sleep have explored \cite{gonzalez_can_2020, norman_methods_2005, singh_model_2022}.

\begin{figure}[!ht]
\centering
\includegraphics[scale=0.28]{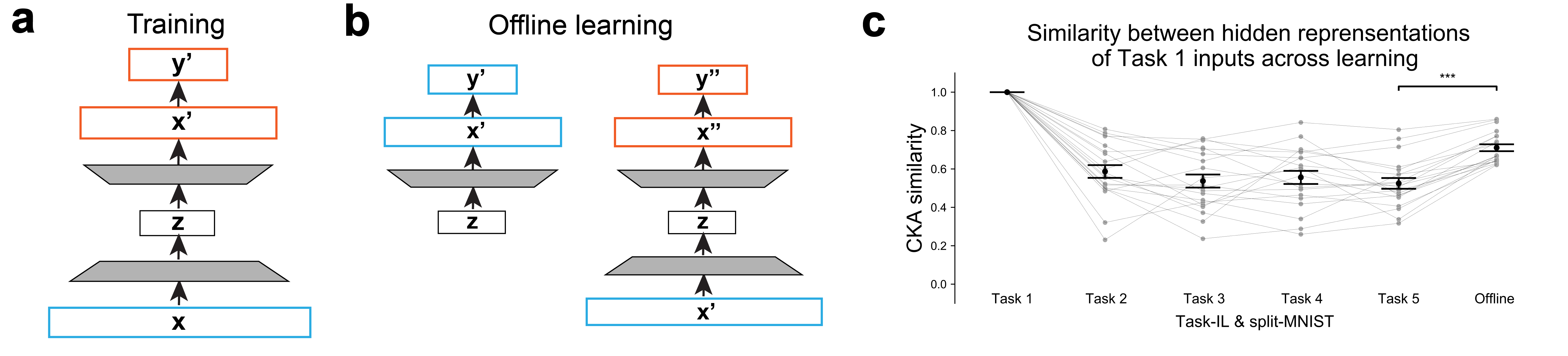}
% \captionsetup{width=0.85\linewidth,font=footnotesize}
\caption{The architecture of our model and its offline recovery of hidden representations. \textbf{a}. We implemented the model as a variational autoencoder with feedforward connections that map the reconstruction layer to a classification output. Each shaded trapezoid represents a stack of hidden layers. During training, the model receives input-output pairs $(x, y)$ and is trained to output $(x', y')$ that approximate $(x, y)$. For CIFAR-100 experiments, the model (not illustrated here) learns to output $(h', y')$ that approximate $(h, y)$, where $h$ is a hidden representation of $x$. \textbf{b}. After the model is trained on all available tasks, it learns from its own replayed samples offline. The model first generates $(x', y')$ pairs as offline training data (left). Given $x'$, the model outputs $(x'', y'')$ and learns to minimize the discrepancy between $(x', y')$ and $(x'', y'')$. In CIFAR-100 experiments, the model (not illustrated here) minimizes the difference between $(h', y')$ and $(h'', y'')$, which are, respectively, the model's replayed samples and the model's outputs given replayed hidden representations $h'$ as inputs. \textbf{c}. Analyses of the similarity between hidden layers' representations of Task 1 inputs across stages of learning suggest that the model can recover prior hidden representations, even those corresponding to the very first task learned, through offline replay. The figure shows an example of this analysis for a hidden layer — the layer that sits directly on top of the latent layer z.
}
\label{fig:examples}
\end{figure}

In contrast, standard continual learning evaluations in machine learning settings do not assess offline algorithms' ability to re-organize memories in an offline period: The effectiveness of algorithms in these settings is quantified by how well they help retain past memories as ANNs sequentially learn multiple tasks (Fig. 1a). Therefore, it is unclear if these algorithms can benefit behavior in the same way the offline brain does. We highlight that, although generative replay effectively reduces forgetting in such settings and has been proposed to mimic biological replay, it affords no learning when trained on its own replay samples (Fig. 1c). Standard generative replay \cite{shin_continual_2017} consists of two components: a generator that learns to reconstruct inputs $x$ and a solver that learns to map inputs to their target outputs $y$. To replay input-output pairs $(x', y')$ that reflect the model's knowledge of past tasks, the generator samples $x'$ that reflect its learned reconstruction of past inputs and the solver labels $x'$ as $y'$ according to its learned mapping. When mixed with data from a new task, $(x', y')$ can serve to preserve the knowledge of past tasks. However, when trained exclusively on $(x', y')$, the model shows no change in behavior, because $(x', y')$ is fully consistent with the solver's learned input-output mapping. As a result, standard generative replay does not self-reorganize memories offline.

\section{An architecture that self-recovers memory}

In this section, we propose a novel architecture that augments generative replay with a brain-like capacity to reorganize memories offline: The ability to self-recover damaged memories. We assess this capacity of the model using two continual learning benchmarks: split-MNIST \cite{zenke_continual_2017} and a split CIFAR-100 task. After the model learns all tasks, we include an offline period where the model learns exclusively from its own generated replay samples. To assess autonomous offline learning, we measure the change in the model's performance on testing data before and after this offline period. 

\subsection{The architecture of the model}

Our proposed model (Fig. 2a) is a generative model trained to concurrently reconstruct a representation of each input $x$ and produce a target output $y$. We implemented the model as a variational autoencoder with feedforward connections mapping the reconstruction layer (i.e., the layer that outputs the reconstructed representation of an input $x$) to an output layer of the same dimension as $y$ for classification. A neurally-plausible feature of the architecture is that, unlike standard generative replay, the model does not employ separate pathways for generating input patterns and labels. As in prior work \cite{van_de_ven_brain-inspired_2020}, we trained the model to minimize the sum of a generative loss (i.e., the sum of a reconstruction loss and a variational loss) and a cross-entropy classification loss. During offline replay, the model first samples $(x', y')$ pairs via the decoder. Generated samples $x'$ then go through a full forward pass through the model, producing the model's reconstruction and classification of $x'$: $(x'', y'')$. For offline learning, we consider $(x', y')$ as the target outputs and $(x'', y'')$ as the model's outputs (Fig. 2b). The offline model is trained to minimize the sum of a generative loss and a distillation loss (see section A.2) computed with respect to $(x', y')$ and $(x'', y'')$.  

\subsection{Experiments on MNIST}

In the first set of experiments, we tested the model on the split-MNIST task, in which the the MNIST dataset is divided into five two-way classification tasks. During training, the model sequentially learned the five tasks one-at-a-time. For training, we considered three continual learning protocols that vary in terms of whether the task identity is provided at the time of test (see section A.4): task-incremental (i.e., Task-IL), domain-incremental (i.e., Domain-IL), and class-incremental learning (i.e., Class-IL) \cite{van_de_ven_three_2019}. After learning, the model learns from its own replayed samples offline. Our main interest was whether the model would self-improve performance through this offline period. To allow room for performance improvement, we omitted replay in between tasks in the Task-IL and Domain-IL scenarios to avoid ceiling performance prior to the offline period.    

To measure change in performance across the offline period, we tested the model on the same set of held-out testing data both prior to and subsequent to the offline period. Across all scenarios and runs of the model initialized with different random seeds, performance improves through the offline period (Table 1). In contrast, neither standard generative replay nor a recently proposed modification of generative replay, BI-R, \cite{van_de_ven_brain-inspired_2020} show reliable performance improvement through offline replay (Table 1). We performed a control experiment by randomly shuffling the pairings of $x'$ and $y'$ in the replayed data. In this control experiment, we observed impaired performance through the offline period (Table 1), suggesting that the pairings of of $x'$ and $y'$ were essential to the model's offline self-recovery of memory. In sum, our results demonstrate that the model self-recovers impaired memories through offline replay. 

\subsection{Tracking changes in hidden representations across tasks and offline learning}

To gain insight into how offline replay promotes memory recovery in the model, we measured the change in the model's hidden representations across tasks and offline learning in split-MNIST experiments. For this analysis, we withheld a set of inputs from Task 1. We computed the similarity between each layer's representations of these inputs at different stages of learning using centered kernel alignment (CKA) \cite{kornblith_similarity_2019, ramasesh_anatomy_2020}. We observed that offline learning facilitated the recovery of all layers' initial representations of Task 1 inputs (i.e., each layer's hidden representation at the end of Task 1 learning): Across all layers, initial representations are more similar to post-offline representations than to pre-offline representations (i.e., p<0.05 for post- vs. pre- offline improvement in CKA similarity across all layers in all three scenarios; an example is shown in Fig. 2c).

\begin{table}
  \caption{Mean post- vs. pre- offline accuracy change. For each experiment, each model was run 10 times with different random seeds, with mean ($\pm$ SEM) accuracy across runs shown below.}
  \label{sample-table}
  \centering
  \scalebox{0.85}{
  \begin{tabular}{lllll}
    \toprule
    % \multicolumn{3}{c}{Part}                   \\
    % \cmidrule(r){1-2}
     & Dataset & Task-IL & Domain-IL & Class-IL \\
    \midrule
    Our model &  split-MNIST & 3.24 ($\pm 0.42$)  & 23.30 ($\pm 0.57$) & 23.91 ($\pm 1.28$) \\
    \midrule
    BI-R &  split-MNIST & -1.50 ($\pm 1.12$)  & 0.13 ($\pm 0.06$)  &  1.55 ($\pm 0.22$) \\
    \midrule
    Standard generative replay & split-MNIST & 0.00 ($\pm 0.00$) & 0.00 ($\pm 0.00$) & 0.00 ($\pm 0.00$) \\
    \midrule
    Control with shuffled targets & split-MNIST & -32.71 ($\pm 1.17$) & -8.68 ($\pm 1.84$) & -26.48 ($\pm 0.89$) \\
    \midrule
    Our model &  CIFAR-100 & 3.30 ($\pm 0.33$)  & 6.55 ($\pm 0.24$) & 8.59 ($\pm 0.38$) \\
    \bottomrule
  \end{tabular}}
\end{table}

\subsection{Extending the model to CIFAR-100}

To identify whether the memory recovery capacity of the model extends to more challenging settings, we tested the model on the CIFAR-100 dataset consisting of naturalistic image categories. We trained the model to sequentially learn 10 tasks, each of which is a ten-way classification task with ten image classes from CIFAR-100. Due to the difficulty of CIFAR-100, for simulations with this dataset, we adapted BI-R, a modification of generative replay that shows robust learning on this task \cite{van_de_ven_brain-inspired_2020}. As in the model we used for MNIST experiments, we added feedforward connections on top of the reconstruction layer for mapping reconstructed representations to classification outputs. Similar to our experiments on MNIST, we disabled replay in the Task-IL and Domain-IL scenarios, and trained the model on its replayed samples after all task learning. On CIFAR-100, we again observed improved performance through the offline period across training scenarios. These results suggest that the memory-recovery capacity of the model generalizes to more complex tasks.

\section{Conclusion}

In the brain, offline replay is generative and appears to benefit memories in the absence of external data. In contrast, standard generative replay in artificial systems does not facilitate performance when trained on its internally-generated replay data. We introduce an architecture that endows generative replay with a capacity of self-reorganization: Self-recovery of damaged memories. The architecture is a generative model augmented with feedforward connections for generating labels. After sequentially learning a set of tasks on split-MNIST and CIFAR-100, our proposed architecture self-recovers damaged memories when trained on its internally-generated data. By contrast, we do not observe this capacity in standard generative replay, in a recent extension of generative replay, nor in our model when its replayed labels are shuffled. These results suggest that including task labels as part of the generative process could enable generative replay models to learn from and self-reorganize existing representations.

% \section*{References}

% \small

\vspace*{1.5cm}

\bibliographystyle{plain}
\bibliography{bib_generative}

\begin{thebibliography}{10}

\bibitem{baran_rem-dependent_2010}
Bengi Baran, Jessica Wilson, and Rebecca Spencer.
\newblock {REM}-dependent repair of competitive memory suppression.
\newblock {\em Experimental brain research}, 203(2):471--477, 2010.

\bibitem{carr_hippocampal_2011}
Margaret~F. Carr, Shantanu~P. Jadhav, and Loren~M. Frank.
\newblock Hippocampal replay in the awake state: a potential substrate for
  memory consolidation and retrieval.
\newblock {\em Nature neuroscience}, 14(2):147--153, 2011.

\bibitem{chaudhry_tiny_2019}
Arslan Chaudhry, Marcus Rohrbach, Mohamed Elhoseiny, Thalaiyasingam Ajanthan,
  Puneet~K. Dokania, Philip~HS Torr, and Marc'Aurelio Ranzato.
\newblock On tiny episodic memories in continual learning.
\newblock {\em arXiv preprint arXiv:1902.10486}, 2019.

\bibitem{drosopoulos_sleeps_2007}
Spyridon Drosopoulos, Claudia Schulze, Stefan Fischer, and Jan Born.
\newblock Sleep's function in the spontaneous recovery and consolidation of
  memories.
\newblock {\em Journal of Experimental Psychology: General}, 136(2):169, 2007.

\bibitem{foster_replay_2017}
David~J. Foster.
\newblock Replay comes of age.
\newblock {\em Annu. Rev. Neurosci}, 40(581-602):9, 2017.

\bibitem{french_catastrophic_1999}
Robert~M. French.
\newblock Catastrophic forgetting in connectionist networks.
\newblock {\em Trends in cognitive sciences}, 3(4):128--135, 1999.

\bibitem{gonzalez_can_2020}
Oscar~C. González, Yury Sokolov, Giri~P. Krishnan, Jean~Erik Delanois, and
  Maxim Bazhenov.
\newblock Can sleep protect memories from catastrophic forgetting?
\newblock {\em Elife}, 9, 2020.

\bibitem{hayes_replay_2021}
Tyler~L. Hayes, Giri~P. Krishnan, Maxim Bazhenov, Hava~T. Siegelmann,
  Terrence~J. Sejnowski, and Christopher Kanan.
\newblock Replay in deep learning: {Current} approaches and missing biological
  elements.
\newblock {\em Neural Computation}, 33(11):2908--2950, 2021.

\bibitem{kemker_fearnet_2017}
Ronald Kemker and Christopher Kanan.
\newblock Fearnet: {Brain}-inspired model for incremental learning.
\newblock {\em arXiv preprint arXiv:1711.10563}, 2017.

\bibitem{kornblith_similarity_2019}
Simon Kornblith, Mohammad Norouzi, Honglak Lee, and Geoffrey Hinton.
\newblock Similarity of neural network representations revisited.
\newblock In {\em International {Conference} on {Machine} {Learning}}, pages
  3519--3529. PMLR, 2019.

\bibitem{krause_large_2022}
Emma~L. Krause and Jan Drugowitsch.
\newblock A large majority of awake hippocampal sharp-wave ripples feature
  spatial trajectories with momentum.
\newblock {\em Neuron}, 110(4):722--733, 2022.

\bibitem{krizhevsky_learning_2009}
Alex Krizhevsky and Geoffrey Hinton.
\newblock Learning multiple layers of features from tiny images.
\newblock 2009.

\bibitem{landmann_reorganisation_2014}
Nina Landmann, Marion Kuhn, Hannah Piosczyk, Bernd Feige, Chiara Baglioni, Kai
  Spiegelhalder, Lukas Frase, Dieter Riemann, Annette Sterr, and Christoph
  Nissen.
\newblock The reorganisation of memory during sleep.
\newblock {\em Sleep medicine reviews}, 18(6):531--541, 2014.

\bibitem{lecun_gradient-based_1998}
Yann LeCun, Léon Bottou, Yoshua Bengio, and Patrick Haffner.
\newblock Gradient-based learning applied to document recognition.
\newblock {\em Proceedings of the IEEE}, 86(11):2278--2324, 1998.

\bibitem{li_learning_2017}
Zhizhong Li and Derek Hoiem.
\newblock Learning without forgetting.
\newblock {\em IEEE transactions on pattern analysis and machine intelligence},
  40(12):2935--2947, 2017.

\bibitem{liu_decoding_2022}
Yunzhe Liu, Matthew~M. Nour, Nicolas~W. Schuck, Timothy~EJ Behrens, and
  Raymond~J. Dolan.
\newblock Decoding cognition from spontaneous neural activity.
\newblock {\em Nature Reviews Neuroscience}, 23(4):204--214, 2022.

\bibitem{mcclelland_why_1995}
James~L. McClelland, Bruce~L. McNaughton, and Randall~C. O'Reilly.
\newblock Why there are complementary learning systems in the hippocampus and
  neocortex: insights from the successes and failures of connectionist models
  of learning and memory.
\newblock {\em Psychological review}, 102(3):419, 1995.

\bibitem{mccloskey_catastrophic_1989}
Michael McCloskey and Neal~J. Cohen.
\newblock Catastrophic interference in connectionist networks: {The} sequential
  learning problem.
\newblock In {\em Psychology of learning and motivation}, volume~24, pages
  109--165. Elsevier, 1989.

\bibitem{mcdevitt_rem_2015}
Elizabeth~A. McDevitt, Katherine~A. Duggan, and Sara~C. Mednick.
\newblock {REM} sleep rescues learning from interference.
\newblock {\em Neurobiology of learning and memory}, 122:51--62, 2015.

\bibitem{mednick_restorative_2002}
Sara~C. Mednick, Ken Nakayama, Jose~L. Cantero, Mercedes Atienza, Alicia~A.
  Levin, Neha Pathak, and Robert Stickgold.
\newblock The restorative effect of naps on perceptual deterioration.
\newblock {\em Nature neuroscience}, 5(7):677--681, 2002.

\bibitem{mnih_human-level_2015}
Volodymyr Mnih, Koray Kavukcuoglu, David Silver, Andrei~A. Rusu, Joel Veness,
  Marc~G. Bellemare, Alex Graves, Martin Riedmiller, Andreas~K. Fidjeland, and
  Georg Ostrovski.
\newblock Human-level control through deep reinforcement learning.
\newblock {\em Nature}, 518(7540):529--533, 2015.

\bibitem{norman_methods_2005}
Kenneth~A. Norman, Ehren~L. Newman, and Adler~J. Perotte.
\newblock Methods for reducing interference in the complementary learning
  systems model: oscillating inhibition and autonomous memory rehearsal.
\newblock {\em Neural Networks}, 18(9):1212--1228, 2005.

\bibitem{ramasesh_anatomy_2020}
Vinay~V. Ramasesh, Ethan Dyer, and Maithra Raghu.
\newblock Anatomy of catastrophic forgetting: {Hidden} representations and task
  semantics.
\newblock {\em arXiv preprint arXiv:2007.07400}, 2020.

\bibitem{rasch_about_2013}
Björn Rasch and Jan Born.
\newblock About sleep's role in memory.
\newblock {\em Physiological reviews}, 2013.

\bibitem{ratcliff_connectionist_1990}
Roger Ratcliff.
\newblock Connectionist models of recognition memory: constraints imposed by
  learning and forgetting functions.
\newblock {\em Psychological review}, 97(2):285, 1990.

\bibitem{rebuffi_icarl_2017}
Sylvestre-Alvise Rebuffi, Alexander Kolesnikov, Georg Sperl, and Christoph~H.
  Lampert.
\newblock icarl: {Incremental} classifier and representation learning.
\newblock In {\em Proceedings of the {IEEE} conference on {Computer} {Vision}
  and {Pattern} {Recognition}}, pages 2001--2010, 2017.

\bibitem{robins_catastrophic_1995}
Anthony Robins.
\newblock Catastrophic forgetting, rehearsal and pseudorehearsal.
\newblock {\em Connection Science}, 7(2):123--146, 1995.

\bibitem{rolnick_experience_2019}
David Rolnick, Arun Ahuja, Jonathan Schwarz, Timothy Lillicrap, and Gregory
  Wayne.
\newblock Experience replay for continual learning.
\newblock {\em Advances in Neural Information Processing Systems}, 32, 2019.

\bibitem{schwarz_progress_2018}
Jonathan Schwarz, Wojciech Czarnecki, Jelena Luketina, Agnieszka
  Grabska-Barwinska, Yee~Whye Teh, Razvan Pascanu, and Raia Hadsell.
\newblock Progress \& compress: {A} scalable framework for continual learning.
\newblock In {\em International {Conference} on {Machine} {Learning}}, pages
  4528--4537. PMLR, 2018.

\bibitem{shin_continual_2017}
Hanul Shin, Jung~Kwon Lee, Jaehong Kim, and Jiwon Kim.
\newblock Continual learning with deep generative replay.
\newblock {\em Advances in neural information processing systems}, 30, 2017.

\bibitem{singh_model_2022}
Dhairyya Singh, Kenneth~A. Norman, and Anna~C. Schapiro.
\newblock A model of autonomous interactions between hippocampus and neocortex
  driving sleep-dependent memory consolidation.
\newblock {\em Proceedings of the National Academy of Sciences of the United
  States of America}, 119(44), 2022.

\bibitem{stella_hippocampal_2019}
Federico Stella, Peter Baracskay, Joseph O’Neill, and Jozsef Csicsvari.
\newblock Hippocampal reactivation of random trajectories resembling {Brownian}
  diffusion.
\newblock {\em Neuron}, 102(2):450--461, 2019.

\bibitem{stickgold_sleep-dependent_2005}
Robert Stickgold.
\newblock Sleep-dependent memory consolidation.
\newblock {\em Nature}, 437(7063):1272--1278, 2005.

\bibitem{van_de_ven_brain-inspired_2020}
Gido~M. van~de Ven, Hava~T. Siegelmann, and Andreas~S. Tolias.
\newblock Brain-inspired replay for continual learning with artificial neural
  networks.
\newblock {\em Nature communications}, 11(1):1--14, 2020.

\bibitem{van_de_ven_three_2019}
Gido~M. van~de Ven and Andreas~S. Tolias.
\newblock Three scenarios for continual learning.
\newblock {\em arXiv preprint arXiv:1904.07734}, 2019.

\bibitem{zenke_continual_2017}
Friedemann Zenke, Ben Poole, and Surya Ganguli.
\newblock Continual learning through synaptic intelligence.
\newblock In {\em International {Conference} on {Machine} {Learning}}, pages
  3987--3995. PMLR, 2017.

\end{thebibliography}

\newpage

%%%%%%%%%%%%%%%%%%%%%%%%%%%%%%%%%%%%%%%%%%%%%%%%%%%%%%%%%%%%

\section*{A. Supplementary information}
\subsection*{A.1 Network architecture}

The architecture of our proposed model is a variational autoencoder with feedforward connections for classification. The model consists of (i) an encoder network $e_\phi$ that maps an input vector $x$ to a vector of latent variables $z$, for which each unit consists of two values that parameterize a Gaussian distribution (i.e., the two values of each unit respectively represent the mean and standard deviation of a Gaussian distribution), (ii) a decoder network $d_\psi$ that learns to map $z$ to a vector $x'$ that approximates inputs $x$, and (iii) a feedfoward classifier $c_{\omega}$ that maps $x'$ to an output $y'$ for classification. Model parameters of $e_\phi$, $d_\psi$, and $c_{\omega}$ closely follow those reported for BI-R \cite{van_de_ven_brain-inspired_2020}. In MNIST experiments, $e_\phi$ and $d_\psi$ are both fully connected networks that each consist of two hidden layers with 400 hidden units, with ReLU activation functions. In CIFAR experiments, $e_\phi$ consists of five pre-trained convolutional layers, which are identical to those employed in BI-R \cite{van_de_ven_brain-inspired_2020}, and two fully connected layers with 2000 hidden ReLU units, whereas $d_\psi$ contains two fully connected layers with 2000 hidden ReLU units. In both experiments, the architectures of $c_{\omega}$, standard generative replay, and BI-R are identical to the those employed in van de Ven et al. \cite{van_de_ven_brain-inspired_2020}. \\

The architecture of our proposed model differs from that of standard generative replay and BI-R. In standard generative replay, the generative model (i.e., the generator) and the classifier do not share any connections and are trained separately (i.e., the generative model is trained on images, the classifier is trained on labels). In contrast, our model combines the generator (i.e., the variational autoencoder in our implementation) and the classifier into a single model. During training on external inputs $x$, the classifier takes in $x'$ — the generator's reconstruction of $x$, as input. In our model, the classifier takes the output of the decoder as input, which allows task labels to be part of the generative (i.e., decoding) process. In contrast, in BI-R, the classifier takes in a hidden representation within the encoder as input.

\subsection*{A.2 Loss function}

The model learns to minimize a loss $L_{total}$: 
\begin{gather*}
L_{total} = L_{r}+L_{v}+L_{c}
\end{gather*}
where $L_{\text{r}}$ is a reconstruction loss, $L_{\text{v}}$ is a variational loss, and $L_{\text{c}}$ is a classification loss .

$L_{\text{v}}$, which measures the extent to which Gaussian distributions specified by outputs of the encoder deviate from standard normal distributions, has the following definition for each sample $\textbf{x}$:
\begin{gather*}
L_\text{v}(\textbf{x};\phi) = \frac{1}{2} \sum_{i=1}^D (1+\log (\sigma_j^{(\textbf{x})^2})-\mu_j^{(\textbf{x})^2}-\sigma_j^{(\textbf{x})^2})
\end{gather*}
where $\mu_j^{(\textbf{x})}$ and $\sigma_j^{(\textbf{x})}$ are respectively the $j^\text{th}$ elements of $\mu^{(\textbf{x})}$ and $\sigma^{(\textbf{x})}$, which are the outputs of the encoder network $e_\phi$ given input $\textbf{x}$, and $D$ is the length of the latent vector $z$.

$L_{\text{r}}$, which quantifies the discrepancy between input $\textbf{x}$ and the output of the decoder network $d_\psi$, is defined as:
\begin{gather*}
L_\text{r}(\textbf{x};\phi,\psi) = \sum_{i=1}^{N}x_i \log ({\textbf{x'}}_i) + (1-\textbf{x}_i)\log (1-{\textbf{x'}}_i)
\end{gather*}
where $\textbf{x}_i$ is the value of the $i^\text{th}$ unit of the input $\textbf{x}$, $\textbf{x'}_i$ is the activation of the $i^\text{th}$ unit of the decoder's output $\textbf{x'}=d_\psi(z^{(x)})$ with $z^{(\textbf{x})}$ = $\mu^{(\textbf{x})}+\sigma^{(\textbf{x})}*\epsilon$ whereby $\epsilon$ is sampled from $X \sim \mathcal{N}(0,I)$, and $N$ is the length of $\textbf{x}$.

$L_{\text{c}}$ measures the discrepancy between the $c_{\omega}$'s output and target output y. When the model is trained on actual tasks, where $y$ is a hard (i.e., one-hot) target, $L_{\text{c}}$ is defined as:
\begin{gather*}
L_\text{c}(\textbf{x}, \textbf{y};\phi,\psi, \omega) = - \log p (Y=\textbf{y}|\textbf{x})
\end{gather*}
where $p$ is conditional probability distribution the network outputs for the input \textbf{x}. In particular, $p$ corresponds to the softmax output of $c_{\omega}$. In the Task-IL but not in the Domain-IL and Class-IL scenarios, the softmax output of $c_{\omega}$ activates only the output units representing classes in either the current task or a replayed task as in prior work \cite{van_de_ven_brain-inspired_2020}.  During training on each replay sample (\textbf{x}', \textbf{y}') where $\textbf{y}'$ serves as a "soft" target with a probability for each active class, $L_{\text{c}}$ is defined as a distillation loss $L_{\text{d}}$: \\
\begin{gather*}
L_\text{d}(\textbf{x}',\textbf{y}';\phi,\psi, \omega)=-{T}^2\sum_{i=1}^{M}\textbf{y}'_i \log p^T (Y=i|\textbf{x}')
\end{gather*}
where $M$ is the number of classes considered and $T$ is a temperature parameter that controls the relative scaling of probabilities that the model outputs. For replay, $\textbf{y}'$ is obtained using a copy of the main model stored after it finishes training on the most recent task.

\subsection*{A.3 Training details}

In our simulations, batch size is 128 for split-MNIST tasks and 256 for split CIFAR tasks. The model undergoes 2000 iterations of training in each split-MNIST task and 5000 training iterations in each split CIFAR task. We used the ADAM-optimizer $(\beta_1 = 0.9, \beta_2 = 0.999)$ with a learning rate of 0.001 in split-MNIST simulations and 0.0001 in split CIFAR simulations. We applied the same number of iterations and batch size during replay.

During online training, the model learns by minimizing $L_{total}$. During offline replay, the generator (i.e., a variational autoencoder) samples $x'$ and the classifier labels $x'$ as $y'$. Given $x'$ as input, the model is trained to minimize $L_{total}$ with $y'$ as the target and $L_c$ as the distillation loss $L_d$. 

Across all simulations, the model learns a sequence of classification tasks, each of which includes a subset of classes from a dataset (i.e., MNIST or CIFAR-100). In split-MNIST, the 10 digit classes are divided into 5 tasks, each of which is a two-way classification task. In comparison, split CIFAR100 consists of 10 tasks, each of which is a ten-way classification task.

In the class scenario but not in the task and domain scenarios, in between tasks, the model generates replay data, which is interleaved with task inputs. In all scenarios, after the presentation of all tasks, in an offline phase, the model learns from replay data generated by a copy of itself. 

\subsection*{A.4 Testing protocol}

Model performance on each task is tested on 8 batches with a batch size of 128 (i.e., 1024 images in total for each task). The exact same testing data is used to evaluate performance of models initialized with different seeds both before and after offline replay on all tasks.

Following prior work \cite{van_de_ven_three_2019}, we assessed the model's performance in three scenarios: Task-IL, Domain-IL, and Class-IL scenarios. The main difference between the three scenarios is the extent to which task identities are available at the time of test. Consider a split-MNIST task in which the model learns five two-way digit classifications tasks sequentially. In the Task-IL scenario, given an input image, the model selects between the two classes from the task that contains the image's actual class. In the Domain-IL scenario, the model selects whether it is the first or second class of a task without knowing the task identity. Finally, in the Class-IL scenario, the model selects from all 10 classes that it has seen so far.  \\

\end{document}